\documentclass[acmtog,screen,nonacm]{acmart}

\usepackage{booktabs} 
\usepackage{bm}
\usepackage{xcolor}
\usepackage{multirow}
\usepackage{array} 
\usepackage{makecell}

\usepackage[ruled]{algorithm2e} 

\SetAlFnt{\small}
\SetAlCapFnt{\small}
\SetAlCapNameFnt{\small}
\SetAlCapHSkip{0pt}

\begin{document}
\title{From Sim-to-Real: Toward General Event-based Low-light Frame Interpolation with Per-scene Optimization }

\author{Ziran Zhang}
\orcid{0000-0001-7003-2735}
\authornote{Equal contribution.}
\email{naturezhanghn@zju.edu.cn}
\affiliation{%
    \institution{Zhejiang University}
        \city{Hangzhou}
        \country{China}
}
\affiliation{
\institution{Shanghai AI Laboratory}
     \city{Shanghai}
     \country{China}
}

\author{Yongrui Ma}
\orcid{0009-0002-7584-0920}
\authornotemark[1]
\email{yongrayma@gmail.com}
\affiliation{%
    \institution{The Chinese University of Hong Kong}
        \city{Hong Kong}
	\country{China}
}
\affiliation{%
    \institution{Shanghai AI Laboratory}
        \city{Shanghai}
	\country{China}
}

\author{Yueting Chen}
\orcid{0000-0002-2759-9784}
\email{chenyt@zju.edu.cn}
\affiliation{%
     \institution{Zhejiang University}
         \city{Hangzhou} 
         \country{China}
}

\author{Feng Zhang}
\orcid{0000-0002-1097-2116}
\email{zhangfeng@pjlab.org.cn}
\affiliation{%
     \institution{Shanghai AI Laboratory}
         \city{Shanghai}
         \country{China}
}

\author{Jinwei Gu}
\orcid{0000-0001-8705-8237}
\email{jwgu@cse.cuhk.edu.hk}
\affiliation{%
     \institution{The Chinese University of Hong Kong}
         \city{Hong Kong}
         \country{China}
}

\author{Tianfan Xue}
\orcid{0000-0001-5031-6618}
\authornote{Corresponding Authors.}
\email{tfxue@ie.cuhk.edu.hk}
\affiliation{%
     \institution{The Chinese University of Hong Kong}
         \city{Hong Kong}
         \country{China}
}

\author{Shi Guo}
\orcid{0000-0001-5155-6162}
\authornotemark[2]
\email{guoshi@pjlab.org.cn}
\affiliation{%
     \institution{Shanghai AI Laboratory}
         \city{Shanghai}
         \country{China}
}

\begin{abstract}

Video Frame Interpolation (VFI) is important for video enhancement, frame rate up-conversion, and slow-motion generation. The introduction of event cameras, which capture per-pixel brightness changes asynchronously, has significantly enhanced VFI capabilities, particularly for high-speed, nonlinear motions. However, these event-based methods encounter challenges in low-light conditions, notably trailing artifacts and signal latency, which hinder their direct applicability and generalization. Addressing these issues, we propose a novel per-scene optimization strategy tailored for low-light conditions. This approach utilizes the internal statistics of a sequence to handle degraded event data under low-light conditions, improving the generalizability to different lighting and camera settings. To evaluate its robustness in low-light condition, we further introduce EVFI-LL, a unique RGB+Event dataset captured under low-light conditions. Our results demonstrate state-of-the-art performance in low-light environments. Project page: \href{https://openimaginglab.github.io/Sim2Real/}{https://openimaginglab.github.io/Sim2Real/}.
\end{abstract}



\begin{teaserfigure}
  \includegraphics[width=\textwidth]{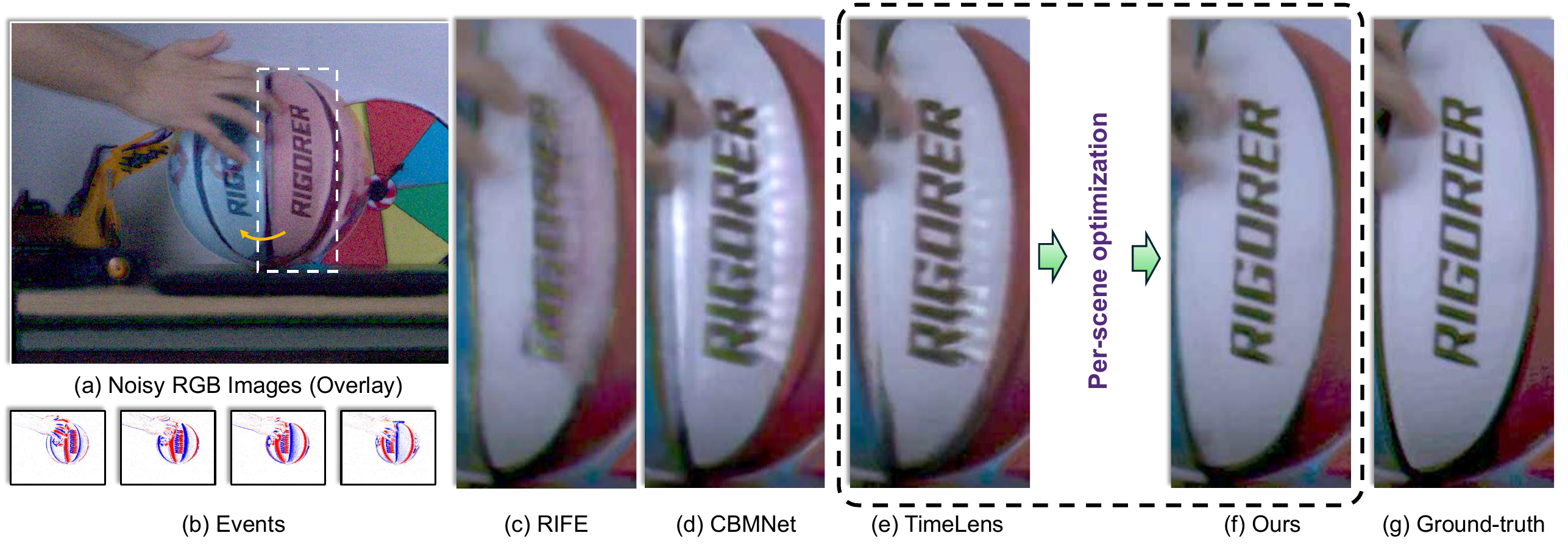}
  \caption{
  The interpolated result from the real-captured RGB-Event sequence under low light conditions. Our proposed per-scene optimization method can successfully correct the impact of event latency, accurately interpolate the correct positions, and produce visually pleasing interpolation results.
  }
  \label{fig:teaser}
\end{teaserfigure}

\maketitle

\section{Introduction}
High-speed imaging is important to computational photography and computer graphics, facilitating the capture of rapidly moving objects and scenes~\cite{yang2022high,zhang2021rosefusion,jiang2018super,choi2020deep}. This is also critical in other complex applications, such as turbulence visualization~\cite{bai2021predicting,wang2024physics}, fast fluorescence lifetime imaging~\cite{lee2019coding}, and high-frequency vibration analysis~\cite{liu2005motion}. Although high-speed cameras are a widely used solution, their high cost and need for intense illumination often restrict their practicality. An effective alternative is video frame interpolation (VFI), which generates high-frame-rate videos from low-frame-rate videos captured with conventional cameras. VFI has been widely applied in video enhancement~\cite{choi2020deep} and slow-motion generation~\cite{jiang2018super}. The advent of event cameras~\cite{lichtsteiner2008128, brandli2014240,guo20233,kodama20231} has demonstrated good potential for VFI, because it can asynchronously capture per-pixel brightness changes with much higher temporal resolution and higher dynamic range.
Recent studies~\cite{tulyakov2021time,tulyakov2022time,sun2023refid,kim2023event} have explored the event-based video frame interpolation (Event-VFI), demonstrating significant performance enhancements compared to the traditional RGB-based VFI~\cite{bao2019depth,sim2021xvfi,huang2022real}.



However, the performance of Event-VFI degrades in low light captures due to the noticeable trailing artifacts associated with event signals~\cite{liu2024seeing, gracca2023shining, graca2021unraveling,hu2021v2e}.
As shown in Fig.~\ref{fig:teaser} (d) and (e), one can observe that in the results from the current Event-VFI methods, such as TimeLens~\cite{tulyakov2021time} and CBMNet~\cite{kim2023event}, there is ghosting of letters due to event latency, and the interpolated frames exhibit noticeable trailing artifacts.
Given that low-light conditions are common, investigating Event-VFI under low-light conditions represents a meaningful, practical, yet challenging task.

The low-light condition presents another challenge to Event-VFI: the latency, noise, and other degradations in the event signals are difficult to model.
One line of solution~\cite{liang2023coherent,zhou2023deblurring,liang2024towards} is to simulate more realistic events by accounting for various degradation factors, including leak noise events, refractory periods, and shot noise, using event simulators, such as v2e~\cite{hu2021v2e} or ESIM~\cite{rebecq2018esim}. However, even with an improved simulation pipeline, a significant discrepancy between simulated data and real-world data still exists~\cite{liu2024seeing}. Another solution~\cite{liu2024seeing} focuses on collecting real-world event datasets to bridge the gap between simulated and actual event signals. However, dataset acquisition is costly, and the datasets tailored to one particular camera with a fixed setting cannot generalize to other cameras or even to other settings. Thus, this raises an interesting and challenging question: \emph{how can the Event-VFI method be cost-effectively generalized to real-world low-light conditions with varying camera settings?}

\begin{figure}[!t]
\includegraphics[width=0.4\textwidth]{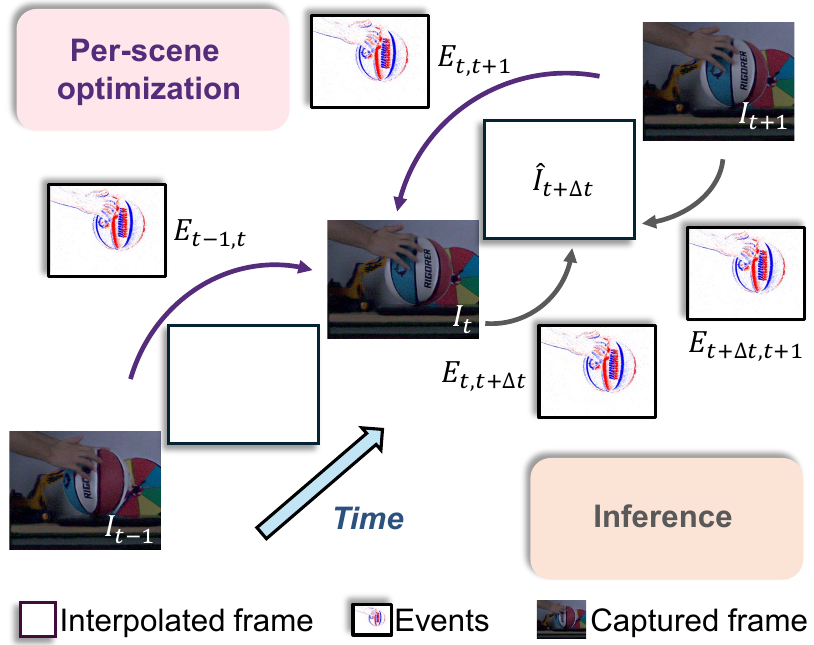}
\caption{Illustration of the proposed per-scene optimization process.}

\label{fig:intro-per-scene}
\end{figure}

To address the above question, we propose a novel per-scene optimization strategy for general Event-VFI in low light conditions. Because the trailing artifacts and other degradations in event signals are hard to model at the training stage, we instead perform per-scene optimization to adapt an interpolation model to a specific distribution of that event stream. This is based on the insight that these degradations are mostly influenced by illuminance, camera hardware, and capture settings, which are relatively constant in a single video.
As shown in Fig.~\ref{fig:intro-per-scene}, unlike previous Event-VFI methods~\cite{tulyakov2021time,tulyakov2022time,kim2023event}, our method first utilizes the entire input sequence to fine-tune a pre-trained Event-VFI model. Specifically, we use the captured RGB image $I_t$ at time $t$ as the ground truth (with denoising applied) to supervise the interpolation results from two other captured RGB frames ($I_{t-1}$ and $I_{t+1}$) at neighboring timestamps and event signals ($E_{t-1,t}$ and $E_{t,t+1}$) during this interval. Then, at the actual inference stage, the fine-tuned network is applied to the same sequence to generate a novel frame $\hat{I}_{t+\Delta t}$ to actually increase the frame rate.

Additionally, we also improve the robustness of pre-trained Event-VFI in low-light conditions. We revisit and refine the low-light event modeling process~\cite{posch2010qvga,liu2024seeing,gracca2023shining,rios2023within,graca2021unraveling}, carefully designing a simulation pipeline tailored for low-light Event-VFI. Utilizing our pre-trained model, our per-scene optimized approach achieves state-of-the-art performance in processing real-world low-light sequences. Additionally, the running time for per-scene optimization is comparable to that of the inference process, thereby ensuring that these improvements are achieved at minimal cost.

To evaluate the proposed solution, we also developed EVFI-LL, a novel RGB-Event interpolation benchmark designed for low-light. Unlike previous datasets captured at normal lighting conditions, such as BS-ERGB~\cite{tulyakov2021time} and ERF-X170FPS~\cite{kim2023event}, EVFI-LL consists of sequences captured under low-light conditions under 35 Lux. Notably, EVFI-LL is the first dataset to capture events using different ON/OFF thresholds (three settings: -20, 0 [default], and 20) on the Prophesee EVK4-HD event sensor. Comprising over 20 testing sequences, this dataset includes accurately synchronized RGB videos and event streams, establishing a new benchmark for evaluating low-light Event-VFI. Additionally, it allows for an assessment of the generalizability of Event-VFI methods across various ON/OFF thresholds. 

\begin{figure*}[!th]
\includegraphics[width=0.95\textwidth]{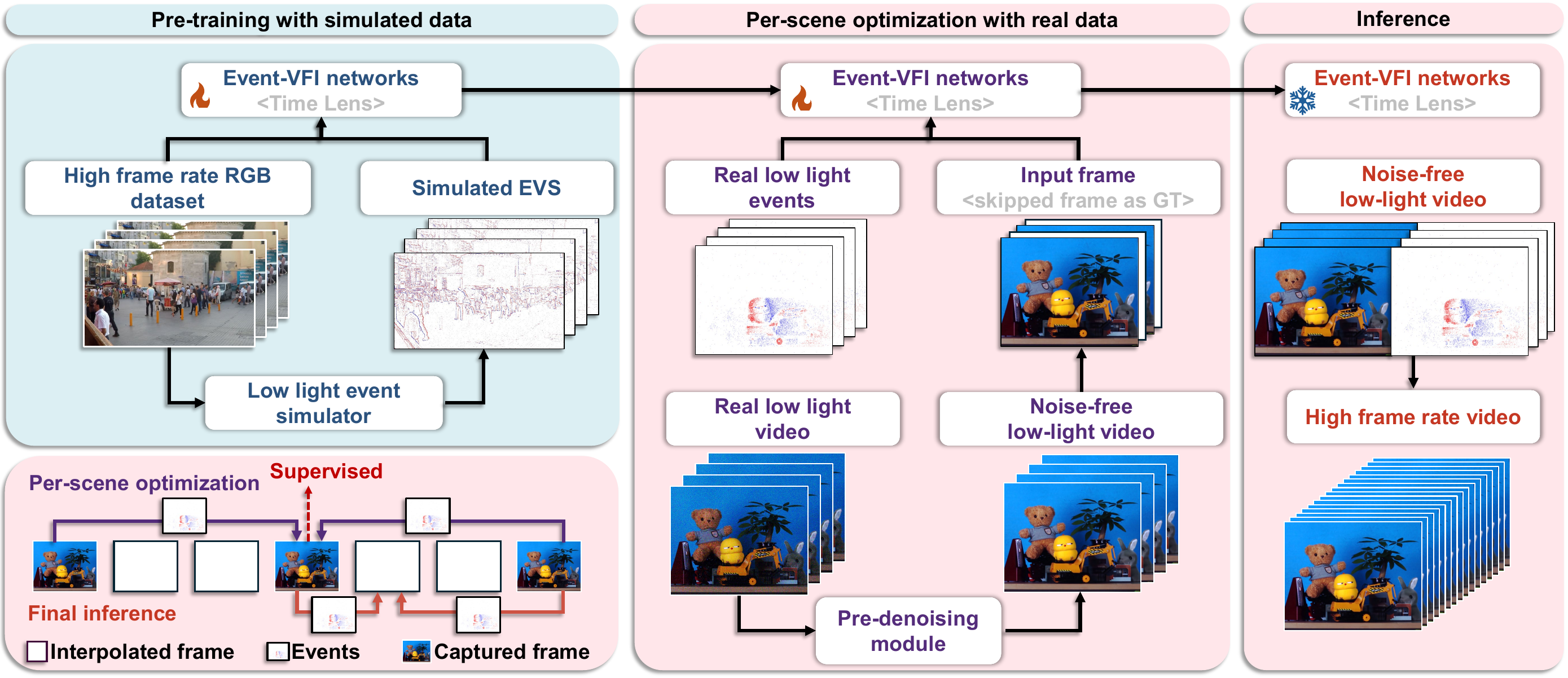}
\caption{Overview of the proposed low light Event-VFI method. This framework contains the pre-training phase, per-scene optimization, and inference process.}
\label{fig:overview}
\end{figure*}

\section{Related Work}
\subsection{RGB-based Video Frame Interpolation}
Video frame interpolation is a fundamental and challenging task that aims to generate inter-frames between consecutive frames in a video. Typical solutions for RGB-based video frame interpolation (RGB-VFI) can be classified into flow-free and flow-based methods. For the flow-free methods, the interpolated frame is generated directly from input frames without explicit flow estimation, with adaptive separable convolution~\cite{niklaus2017video}, phase decomposition~\cite{meyer2018phasenet} and transformer-based structure~\cite{shi2022video}. 
For the Flow-based methods~\cite{bao2019depth,sim2021xvfi,huang2022real,wu2022optimizing}, the optical flow is explicitly estimated in the interpolation. 
However, due to a lack of accurate motion information between two RGB frames, the performance of RGB-VFI degrades when handling complex and large motion.

\subsection{Event-based Video Frame Interpolation}
Event cameras record intensity changes of objects, offering high frame rate and high dynamic range characteristics~\cite{posch2010qvga,chen2024event,gehrig2024low}, which is benefit for VFI task. Thus, using event cameras for VFI (Event-VFI) has recently provided a new direction for addressing non-linear, large motion interpolation~\cite{tulyakov2021time,tulyakov2022time,sun2023refid,kim2023event}. Time Lens~\cite{tulyakov2021time} and Time Lens++~\cite{tulyakov2022time} demonstrated promising performance in non-linear motion scenarios. \citet{sun2023refid} proposed REFID, which jointly performs image deblurring and interpolation. \citet{kim2023event} introduced CBMNet for interpolation with cross-modal asymmetric bidirectional motion field estimation.

Even though Event-VFI methods exhibit better performance in handling non-linear motions, current research has not addressed Event-VFI task in low-light conditions. In low light, event cameras produce noticeable trailing artifacts~\cite{liu2024seeing,posch2010qvga}, which negatively impact event-based interpolation. Thus in this paper, we study this promising yet under-explored low light Event-VFI problem.

\subsection{Low-light Event Correction}
In low light, event cameras produce noticeable trailing artifacts and noise~\cite{liu2024seeing,posch2010qvga,hu2021v2e} which degrade the follow-up tasks. To correct these degradation in low-light events, some methods~\cite{liang2023coherent,zhou2023deblurring} utilize more realistic event simulation. Specially, \citet{liang2023coherent,zhou2023deblurring} utilized the commonly used v2e~\cite{hu2021v2e} or ESIM~\cite{rebecq2018esim} simulator, and considered various degradation factors such as threshold noise, hot pixels, leak noise events, refractory period, and shot noise. However, event signal generated by such simulation method still has great gap with real-captured data and the model trained with simulated data can not directly generalize to real-world events and correct the trailing artifacts of low light events~\cite{liu2024seeing}. 

To correct the real-world latency and noise of event signal, \citet{liu2024seeing} collected a real-world low-light event dataset consisting of low-light events and high-quality images. A learnable event timestamps calibration model is also proposed to learn the event trailing. However, such method needs costly data collection. More importantly, such methods~\cite{liu2024seeing} mainly focusing on correct event signal for specific event camera, specifically for Prophesee EVK4, with fixed camera setting, \emph{e.g.,} ON/OFF thresholds.
However, the hardware differences and parametric setting both effect the latency degree, limiting their generalization. To solve these limitations, in this paper, a per-scene optimization strategy is proposed for low light Event-VFI task to let our model can generalize to event camera with different camera setting without costly data collection.

\section{Problem definition}
\subsection{Low-light Event Representation}
\label{sec:lowlight-event-rep}
Event camera capture scene illumination change as an asynchronous stream of events. Each event at position $\mathbf{u}=(x,y)$ is generated when the relative intensity change is larger than a contrast threshold $c$~\cite{tulyakov2021time,tulyakov2022time,kim2023event}:
\begin{equation}
    E_t = \begin{cases}
        1, & \text{if } \log(I_{t}) - \log(I_{t-\Delta t}) \geq c, \\ 
        -1, & \text{if } \log(I_{t}) - \log(I_{t-\Delta t}) \leq -c,\\        
        0, & \text{otherwise}, 
        \end{cases}       
\label{eq:normal_define_evs}
\end{equation}
where $E_t$ and $I_{t}$ are the event polarity and the image intensity at time $t$, respectively, and $\Delta t$ is the time interval since the last event occurred at $\mathbf{u}$. To simplify the formulation, we omit position $\mathbf{u}$ in the equations above.

\begin{figure}[!t]
\includegraphics[width=\columnwidth]{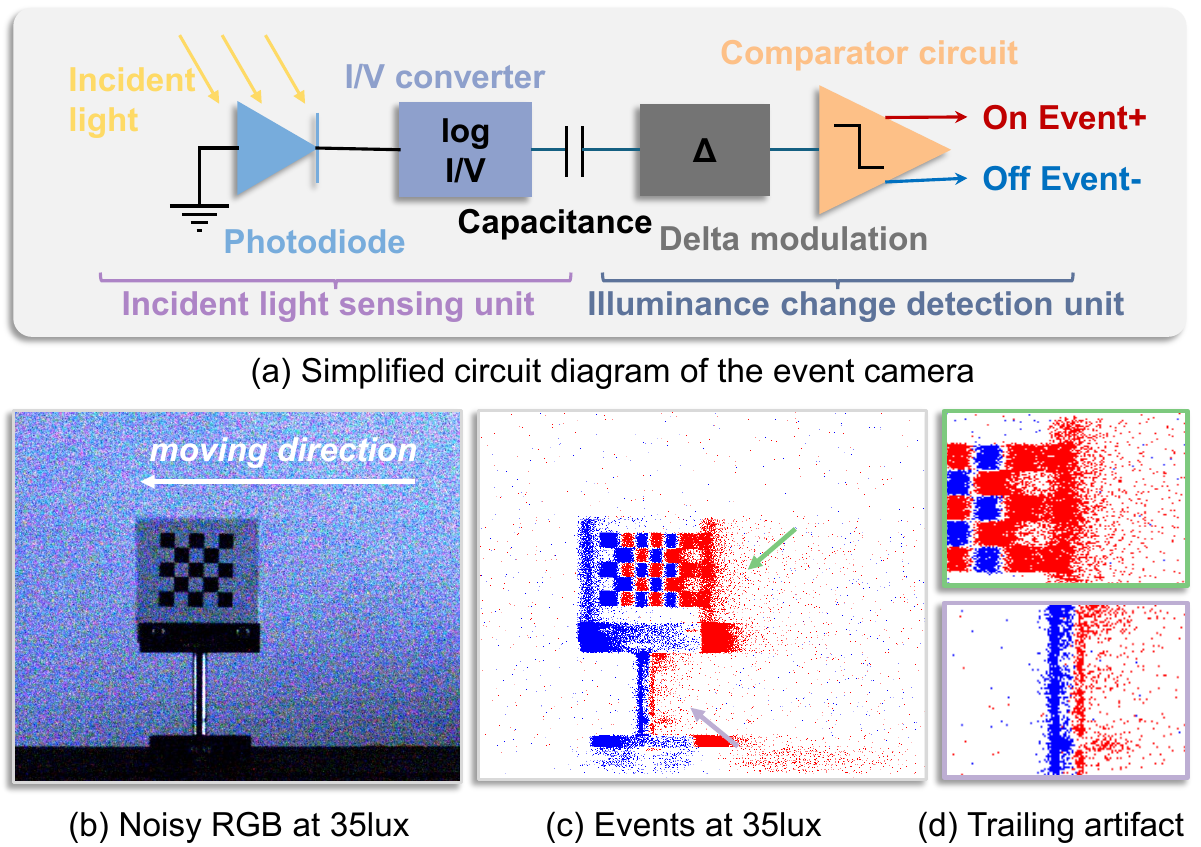}
\caption{Simplified circuit diagram for each pixel of the event camera. And the visualization of event trailing artifacts, captured by our EVS-RGB beam splitting imaging system.}
\label{fig:trail-artifact}
\end{figure}

However, event streams in low light often contain trailing artifacts~\cite{hu2021v2e,liu2024seeing, posch2010qvga} (Fig.~\ref{fig:trail-artifact}(d)), which is not covered by the simplified modeling in Eqn.~\ref{eq:normal_define_evs}. To better model it, we revisit event generation process. As shown in Fig.~\ref{fig:trail-artifact} (a), the photodiode (PD) converts an incoming incident light to photocurrent, whose magnitude is effected by the cutoff frequency $f_c$ of the camera system~\cite{posch2010qvga,liu2024seeing} as
\begin{equation}
    f_{c} = \frac{1}{2\pi C}\frac{I_{ph}}{U_t},
    \label{eqn:cutoff_freq}
\end{equation}
where $C$ is the capacitance, and $U_t$ is the thermal voltage. One can see that $f_c$ is proportional to $I_{ph}$. Thus, the measured intensity or volume for event generation can be modeled as:
\begin{equation}
I'_t = \alpha I_t + (1-\alpha) I_{t-\Delta t},
\label{eq:charge_discharge}
\end{equation}
where $\alpha = 1 - e^{-\Delta t/\tau}$ and $\tau = 1/(2\pi f_c)$. And thus we can approximately obtain the degraded event representation in the low light:
\begin{equation}
    E_t = \begin{cases}
        1, & \text{if } \log(\alpha I_t + (1-\alpha) I_{t-\Delta t}) - \log(I_{t-\Delta t}) \geq c, \\ 
        -1, & \text{if } \log(\alpha I_t + (1-\alpha) I_{t-\Delta t}) - \log(I_{t-\Delta t}) \leq -c,\\        
        0, & \text{otherwise}.
        \end{cases}       
\label{eq:lowlight_define_evs}
\end{equation}

Equations above explain the cause of trailing. When image intensity decreases, the cutoff frequency also decreases (Eqn.~\ref{eqn:cutoff_freq}), leading to a smaller value of $\alpha$. Consequently, the measurement of current intensity is more significantly influenced by prior values, an effect modeled as an RC low-pass filter~\cite{hu2021v2e,liu2024seeing}. Such a configuration implies that the same voltage changes can induce more extended temporal delays in the generation of event signals in the low light condition. In Fig.~\ref{fig:trail-artifact}(d), we display a real-world captured low-light event, where a temporal delay in the event data can be observed, leading to spatial trailing artifacts. Using such degraded event data for frame interpolation results in deformations in the interpolated frames due to the trailing effects (see Fig.~\ref{fig:teaser}).

\subsection{Low Light Event-VFI}
For common Event-VFI, the interpolation is modeled using Eqn.~\ref{eq:normal_define_evs} and formulated as~\cite{jiang2020learning,sun2022event,lin2020learning}:
\begin{equation}
I_{t+\Delta t} \approx I_t \cdot \mbox{exp}\left(\int_t^{t+\Delta t}c\cdot E_{\tau}d\tau\right).
\label{eq:normal-event-vfi}
\end{equation}

In low-light conditions, the event representation defined in Eqn.~\ref{eq:lowlight_define_evs} accounts for temporal latency. When $\Delta t$ is very small, the synthesis frames can be approximately modeled using the degraded event signals:
\begin{equation}
I_{t+\Delta t} \approx \frac{1}{\alpha}(I_t \cdot \mbox{exp}\left(\int_t^{t+\Delta t}c\cdot E_{\tau}d\tau\right) - (1-\alpha)I_t).
\label{eq:ll-event-vfi}
\end{equation}
Compared with Eqn.~\ref{eq:normal-event-vfi}, the low light Event-VFI needs to handle noisy RGB image $I_t$ and consider the impact of trailing artifacts of events which is effect by $\alpha$. 

In recent methods~\cite{tulyakov2021time,tulyakov2022time,kim2023event}, to recover an intermediate RGB frame $I_{t+\Delta t}$ between $I_{t}$ and $I_{t+1}$, information from both the forward (from $I_t$ to $I_{t+\Delta t}$) and the backward (from $I_{t+1}$ to $I_{t+\Delta t}$) estimation are utilized. Thus process of low light Event-VFI in Eqn.~\ref{eq:ll-event-vfi} is solved as:
\begin{equation}
    \hat{I}_{t+\Delta t} =\Phi (I_t, I_{t+1}, E_{t,t+\Delta t},E_{t+\Delta t,t+1}, c, \alpha),
\label{eq:solve-ll-event-vfi}
\end{equation}
where $\Phi$ is the interpolation algorithm.

Note that given that the threshold $c$ and the latency factor $\alpha$, which are influenced by illuminance levels and camera hardware, adapting a pre-trained model to different conditions is challenging. Therefore, in Sec.~\ref{sec:self-learning}, we will propose a per-scene optimization strategy specifically designed for the Event-VFI task, capable of addressing event latency across various camera settings.

\section{Method}
For the low light Event-VFI, the input are noisy low frame rate RGB images $\{I_{t_k}\}_{k=1}^N$ together with the comprehensive event data $E_{t_1\rightarrow t_N}$, where $N$ denotes the sequence length. The objective of Event-VFI is to obtain the interpolated results $I_{t+\Delta t}$ between two neighbor frames $I_t$ and $I_{t+1}$, where $\Delta t \in (t, t+1)$. As illustrated in Eqn.~\ref{eq:solve-ll-event-vfi}, the interpolation in low-light conditions depends not only on the input frames $I_t$ and $I_{t+1}$, events $E_{t,t+1}$, and threshold $c$, but also on the latency or trailing of the event signal $\alpha$ in Eqn.~\ref{eq:charge_discharge}. 

Due to the trailing artifacts and other degradations in events that are difficult to model during training, we will introduce a per-scene optimization method in Sec.~\ref{sec:self-learning}, with its associated loss function presented in Sec.~\ref{sec:loss}. Moreover, to obtain a better per-trained model, we present improvements to the low-light event simulation process in Sec.~\ref{sec:low-light-event-simulation}. The overall pipeline is summarized in Fig.~\ref{fig:overview}.

\subsection{Per-scene Optimization for VFI}
\label{sec:self-learning}
Based on the insight that factors such as illuminance, camera hardware, and capture settings significantly influence event trailing and other degradations, and remain relatively constant within a single video, we propose per-scene optimization. This method allows the Event-VFI backbone to adapt to the current scene's specific event degradation, enabling effective handling of Event VFI under low-light conditions with varying camera settings.

Unlike previous Event-VFI methods~\cite{liang2023coherent,zhou2023deblurring}, which rely solely on two consecutive noisy RGB frames and associated event signals to predict intermediate RGB frames, our per-scene optimization strategy leverages the entire sequence. This approach, depicted in Fig.~\ref{fig:overview}, allows for fine-tuning a pre-trained Event-VFI model using the entire video sequence. It establishes training pairs directly from the testing sequence, using the denoised RGB image $I_t$ at time $t$ as the ground truth to supervise the interpolation results from two other captured RGB frames ($I_{t-n}$ and $I_{t+n}$) at neighboring timestamps, along with event signals ($E_{t-n,t}$ and $E_{t,t+n}$) captured during this interval. Here, $n$ represents the temporal interval between captured RGB frames and is randomly sampled to enrich the data for per-scene optimization. Specifically, in our experiments, the values of $n$ are sampled from 1 to 7 with corresponding probabilities $[0.632, 0.232, 0.086, 0.032, 0.012, 0.004, 0.002]$.

After per-scene fine-tuning, during the actual inference stage, the network utilizes the same sequence to generate a novel frame $\hat{I}_{t+\Delta t}$, effectively increasing the frame rate.

\subsection{Optimization Loss}
\label{sec:loss}
In the per-scene optimization process, the interpolated frame at timestamp $t$, $\hat{I}_t$ can be modeled as:
\begin{equation}
    \hat{I}_t = \Phi(D(I_{t-n}), D(I_{t+n}), E_{t-n, t},E_{t, t+n}),
\end{equation}
where $\Phi(\cdot)$ is the event-based frame interpolation network, $D(\cdot)$ is the real-world denoising operator. In our experiment, we choose state-of-the-art Event-VFI model TimeLens~\cite{tulyakov2021time} as interpolation backbone $\Phi(\cdot)$ and Restormer~\cite{zamir2022restormer} as denoiser $D(\cdot)$. Then the reconstruction loss can be defined as:
\begin{equation}
    \mathcal{L}_r = \sqrt{\Vert \hat{I}_t - D(I_{t}) \Vert^2 + \epsilon^2} + \lambda\sqrt{\Vert S(\hat{I}_t) - S(D(I_{t})) \Vert^2 + \epsilon^2},
\end{equation}
where $\lambda$ is a weighting parameter set to 0.1, $S(\cdot)$ is the Sobel operator,  $\sqrt{\Vert \hat{x} - x \Vert^2 + \epsilon^2}$ is the Charbonnier penalty function, and $\epsilon$ is set to 0.001. Additionally, to obtain visually pleasing interpolation results, the perceptual loss \cite{johnson2016perceptual} is also used:
\begin{equation}
    \mathcal{L}_{p} = \beta \sqrt{\Vert \text{VGG}(\hat{I}_t) - \text{VGG}(D(I_{t})) \Vert^2 + \epsilon^2},
\end{equation}
where $\text{VGG}(\cdot)$ is the pretrained 16 layer VGG network~\cite{russakovsky2015imagenet}, $\beta$ is a weighting parameter which we set to 0.1.

\subsection{Low Light Event Simulation}
\label{sec:low-light-event-simulation}
In this sub-section, we also discuss how to improve the pretrained interpolation model using a more accurate event simulation pipeline. Even with the per-scene optimization proposed above, a robust pretrained model is still critical, as it helps to reduce tuning time and improve performance after per-scene optimization.

\begin{figure*}[h!]
\begin{center}
    \includegraphics[width=0.95\textwidth]{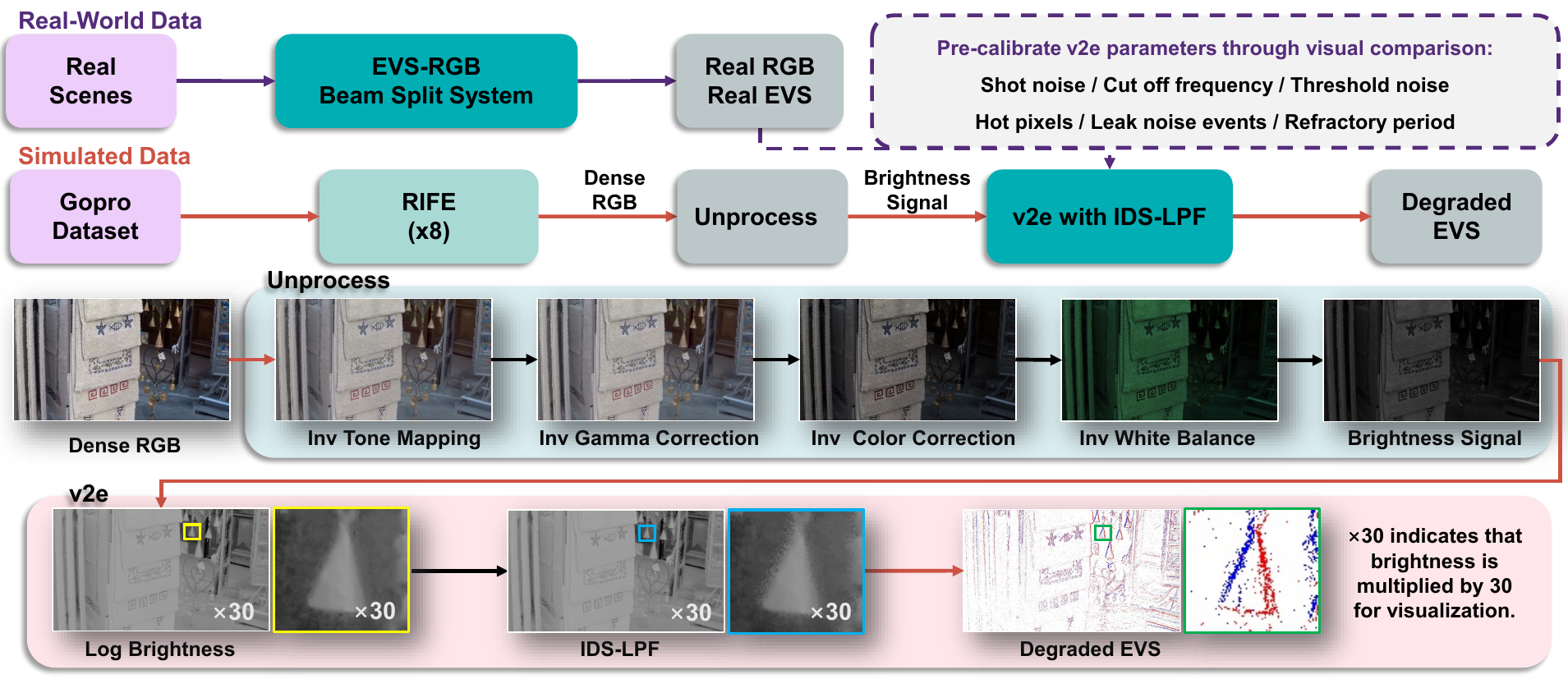}
    \caption{Synthesis process for the EVS Motion Trail Simulation Dataset.}
    \label{fig:simulated_data}
\end{center}
\end{figure*}

Fig.~\ref{fig:simulated_data} shows the proposed simulation pipeline. First, we create high-frame RGB sequences using the GoPro dataset~\cite{zhang2018unreasonable}. Then, we employ RIFE~\cite{huang2022real} for 8$\times$ frame interpolation to generate high frame rate video. Subsequently, we compute a low light intensity image using $L = \Gamma^{-1}(I)$, where $\Gamma^{-1}(\cdot)$ represents the inverse of the image signal processing (ISP) pipeline. Following \cite{brooks2019unprocessing}, $\Gamma^{-1}(\cdot)$ contains inverse tone mapping, inverse gamma correction, inverse color correction, and inverse white balance and brightening to transition the data to a linear brightness domain. 



\begin{figure*}[ht]
\centering
\includegraphics[width=0.95\textwidth]{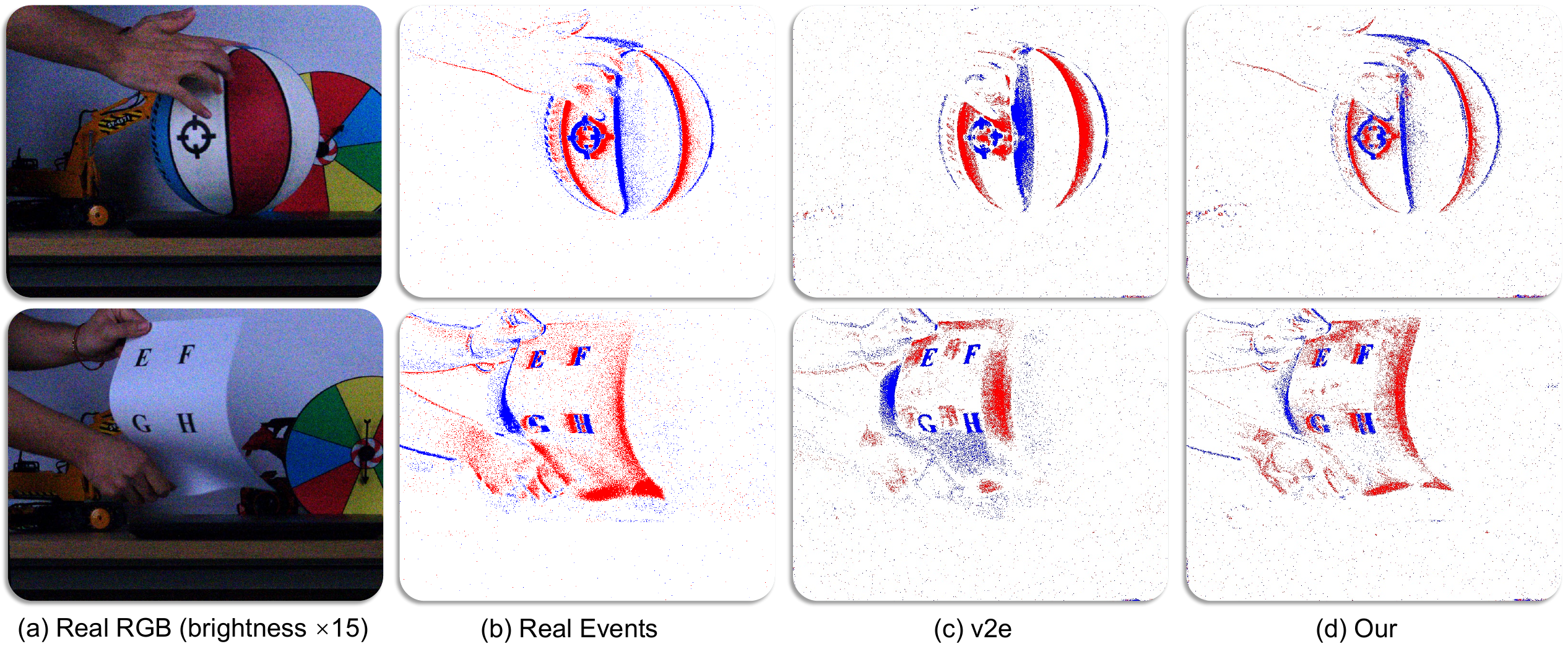}
\caption{Visual comparison of the low-light trailing simulation results of the event camera with v2e.}
\label{fig:tail}
\end{figure*}

Then, we generate events from high-frame-rate linear RGB sequences using the v2e simulator to produce a synthetic EVS-RGB dataset. Under low-light conditions, the performance degradation of event cameras is generally due to motion trails and noise. For the event noise, the v2e simulator~\cite{hu2021v2e} comprehensively models noise factors through parameter adjustments, such as threshold noise, hot pixels, leak noise events, and the refractory period. Before creating simulated data, we calibrated v2e's parameters by comparing simulated data with real-world data. To accurately simulate the motion trailing associated with events, we use a low-pass filter with exponential decay to approximate the RC circuit dynamics described in Eqn.~\ref{eq:charge_discharge} in Sec.~\ref{sec:lowlight-event-rep}. Moreover, although we increase the input video frame rate using the RIFE interpolation method, the resulting video still remains a temporal discretization $\Delta t$ of the event signal. To model temporal signal decay within $\Delta t$ and modeling sparse trailing artifacts, we incorporate a sparse event mask $R$ in the low-pass filter that resets certain event pixels $I_{t-\Delta t}$ to zero. As a result, we reformulate the low-pass filter equation in Eqn.~\ref{eq:charge_discharge} into an Intensity-Dependent Stochastic Low-Pass Filter (IDS-LPF), as shown below:
\begin{equation}
I'_t = R(\alpha, r) I_t + (1-R(\alpha, r)) I_{t-\Delta t},
\label{eq:charge_discharge_random_sample}
\end{equation}
where $R(\alpha, r) = \alpha \cdot \mathbb{I}(r > P) + 1 \cdot \mathbb{I}(r \leq P)$. $r$ is a randomly generated number between 0 and 1, $\mathbb{I}(\cdot)$ is the indicator function that returns 1 if the condition is true and 0 otherwise. In our simulation, $P$ is set to 0.5. Consequently, the indicator function $\mathbb{I}(r > P)$ becomes to 0 when the randomly generated value $r$ is greater than $P$. The visual effects of applying the IDS-LPF have been shown in Fig.~\ref{fig:tail} to better illustrate the process of tailing simulation. The event-RGB data pairs generated from this modified filter are utilized to pretrain the frame interpolation model. By employing this simulation pipeline, our method achieves superior results compared to the original TimeLens, as demonstrated in Tab.~\ref{tab:bia} for low-light Event-VFI.

\begin{table*}[!th]
\centering
\caption{Comparison of different frame interpolation methods on EVFI-LL under various interpolation rates (4$\times$ and 8$\times$) and different bias voltages (Sony IMX636 event sensor, positive and negative threshold voltages with a default bias of 0). The lower the bias, the lower the threshold, the more events recorded.}
\begin{tabular}{@{}llllllll@{}}
\toprule
\multicolumn{1}{c}{Bias Setting} & Method & \multicolumn{3}{c}{4$\times$ Frame Interpolation} & \multicolumn{3}{c}{8$\times$ Frame Interpolation} \\ 
\cmidrule(lr){3-5} \cmidrule(lr){6-8}
 & & \textbf{PSNR $\uparrow$} & \textbf{SSIM $\uparrow$} & \textbf{LPIPS $\downarrow$} & \textbf{PSNR $\uparrow$} & \textbf{SSIM $\uparrow$} & \textbf{LPIPS $\downarrow$} \\

\midrule
\multirow{6}{*}{\centering +20}
 & SuperSloMo~\cite{jiang2018super} & 30.201 & 0.8802 & 0.1371 & 27.665 & 0.8559 & 0.1828  \\
 & RIFE~\cite{huang2022real} & 31.143 & 0.8846 & 0.1420 & 28.523 & 0.8642 & 0.1854 \\
 & TimeLens~\cite{tulyakov2021time} & 30.548 & 0.8673 & 0.1691 & 29.187 & 0.8559 & 0.1887 \\
 & CBMNet~\cite{kim2023event} & 30.114 & 0.8592 & 0.1999 & 28.656 & 0.8465 & 0.2246 \\
 \cmidrule(lr){2-8}
 & Our Pretrained & 30.803 & 0.8805 & 0.1522 & 29.449 & 0.8698 & 0.1708 \\
 & Our Per Scene Opt. & \textbf{31.442} & \textbf{0.8848} & \textbf{0.1285} & \textbf{30.760} & \textbf{0.8795} & \textbf{0.1381} \\

\midrule
\multirow{6}{*}{\centering 0}
 & SuperSloMo~\cite{jiang2018super} & 31.105 & 0.8857 & 0.1264  & 28.724 & 0.8626 & 0.1632 \\
 & RIFE~\cite{huang2022real} & 32.242 & 0.8901 & 0.1301 & 29.462 & 0.8687 & 0.1706 \\
 & TimeLens~\cite{tulyakov2021time} & 30.729 & 0.8647 & 0.1584 & 29.891 & 0.8542 & 0.1728 \\
 & CBMNet~\cite{kim2023event} & 30.323 & 0.8662 & 0.1878 & 29.320 & 0.8552 & 0.2100 \\
  \cmidrule(lr){2-8}
 & Our Pretrained & 31.218 & 0.8836 & 0.1380 & 30.338 & 0.8730 & 0.1505 \\
 & Our Per Scene Opt. & \textbf{32.762} & \textbf{0.8972} & \textbf{0.1172} & \textbf{32.135} & \textbf{0.8879} & \textbf{0.1215}\\

\midrule
\multirow{6}{*}{\centering -20}
 & SuperSloMo~\cite{jiang2018super} & 30.980 & 0.8880 & 0.1269 & 28.912 & 0.8683 & 0.1648 \\
 & RIFE~\cite{huang2022real} & 32.061 & 0.8920 & 0.1307 & 29.702 & 0.8744 & 0.1689 \\
 & TimeLens~\cite{tulyakov2021time} & 31.417 & 0.8743 & 0.1572 & 30.541 & 0.8662 & 0.1706 \\
 & CBMNet~\cite{kim2023event} & 30.838 & 0.8607 & 0.1892 & 29.814 & 0.8509 & 0.2120 \\
\cmidrule(lr){2-8}
 & Our Pretrained & 32.036 & 0.8915 & 0.1329 & 31.241 & 0.8835 & 0.1431 \\
 & Our Per Scene Opt. & \textbf{32.990} & \textbf{0.8977} & \textbf{0.1073} & \textbf{32.529} & \textbf{0.8913} & \textbf{0.1170} \\

\bottomrule
\end{tabular}
\label{tab:bia}
\end{table*}

\section{Experiments}
\subsection{Experiment Details}
We selected the state-of-the-art TimeLens~\cite{tulyakov2021time} as the backbone for our Event-VFI model. Initially, a model tailored for low-light Event-VFI, referred to as Our Pretrained, is trained using simulated data (as detailed in Sec. \ref{sec:low-light-event-simulation}). Before inference phase for each scene, per-scene optimization is first performed using the Adam optimizer~\cite{kingma2014adam} with a learning rate of $1\times 10^{-4}$. The optimized model is denoted as Our Per Scene Opt. It is important to note that to simplify the learning process, we pre-denoise the input RGB images using Restormer~\cite{zamir2022restormer}.

Our method is compared with last RGB-VFI techniques (SuperSloMo~\cite{jiang2018super} and RIFE~\cite{huang2022real}) and state-of-the-art Event-VFI methods (CBMNet~\cite{kim2023event} and TimeLens~\cite{tulyakov2021time}).

\subsection{Low Light Event-VFI Benchmark: EVFI-LL}
Previous real-world Event-VFI datasets, \emph{i.e.}, BS-ERGB~\cite{tulyakov2021time} and ERF-X170FPS~\cite{kim2023event}, were captured under normal lighting conditions and are not suitable for testing the trailing artifacts in low light events. Therefore, we have captured a new low-light Event-VFI benchmark, denoted as EVFI-LL.

\begin{figure}[th]
\begin{center}
    \includegraphics[width=8cm]{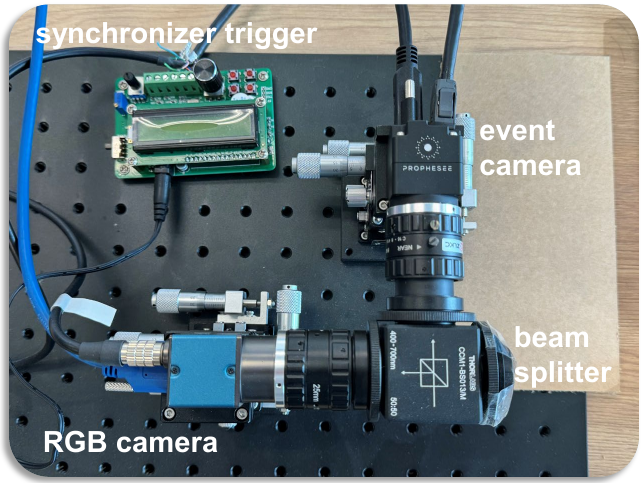}
    \caption{EVS-RGB Beam Splitting Imaging System. Spatial-temporal alignment is ensured through calibration and trigger control.}
    \label{fig:system}
\end{center}
\end{figure}

Similar to ERF-X170FPS~\cite{kim2023event}, our setup employs a beam splitter to align the fields of view of the event and RGB cameras, as depicted in Fig.~\ref{fig:system}. We use a Prophesee EVK4-HD ($1280\times720$) event camera and a MER2-301-125U3C ($2048\times1536$) RGB camera. Our system ensures that both cameras are geometrically calibrated and temporally synchronized. Calibration is achieved by estimating transformations using a blinking checkerboard pattern, and synchronization is managed with synchronization trigger controls.

To capture the low-light Event-RGB dataset, we set the ambient light level to below 35 Lux and used short exposure times to minimize blur. The dataset includes 10 non-linear motion scenarios, such as rotating balls and twisting paper sheets. To test the generalization capability of various methods, we captured additional sequences with varying camera parameters, particularly ON/OFF thresholds. During data acquisition, the RGB camera's frame rate was maintained at 64Hz.

For testing purposes, we subsampled the test video by skipping every N-1 frames, which resulted in an effective testing frame rate reduced by a factor of N. Event data is not skipped during testing. This lower frame rate poses a significant challenge for frame interpolation tasks, using the skipped frames (with denoising applied) as ground truth for performance evaluation. We also curated more challenging scenes with complex and large motions within the EVFI-LL dataset, creating a challenging track denoted as EVFI-LL-C.

\begin{figure*}[ht!]
\begin{center}
    \includegraphics[width=0.93\textwidth]{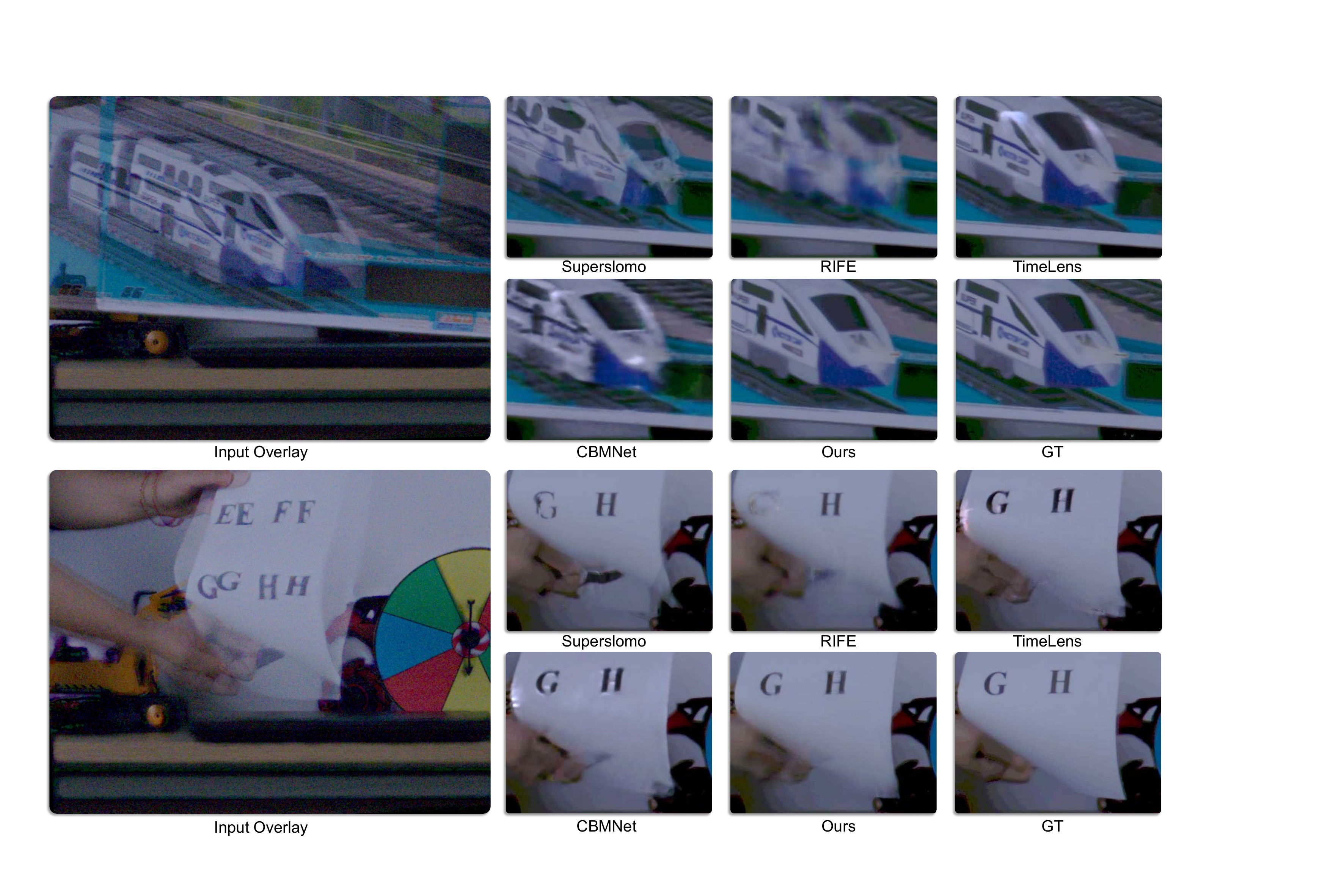}
    \caption{Visual comparison of results from different methods under large motion.}
    \label{fig:large_motion}
\end{center}
\end{figure*}

\begin{figure*}[h]
\centering
\includegraphics[width=0.93\textwidth]{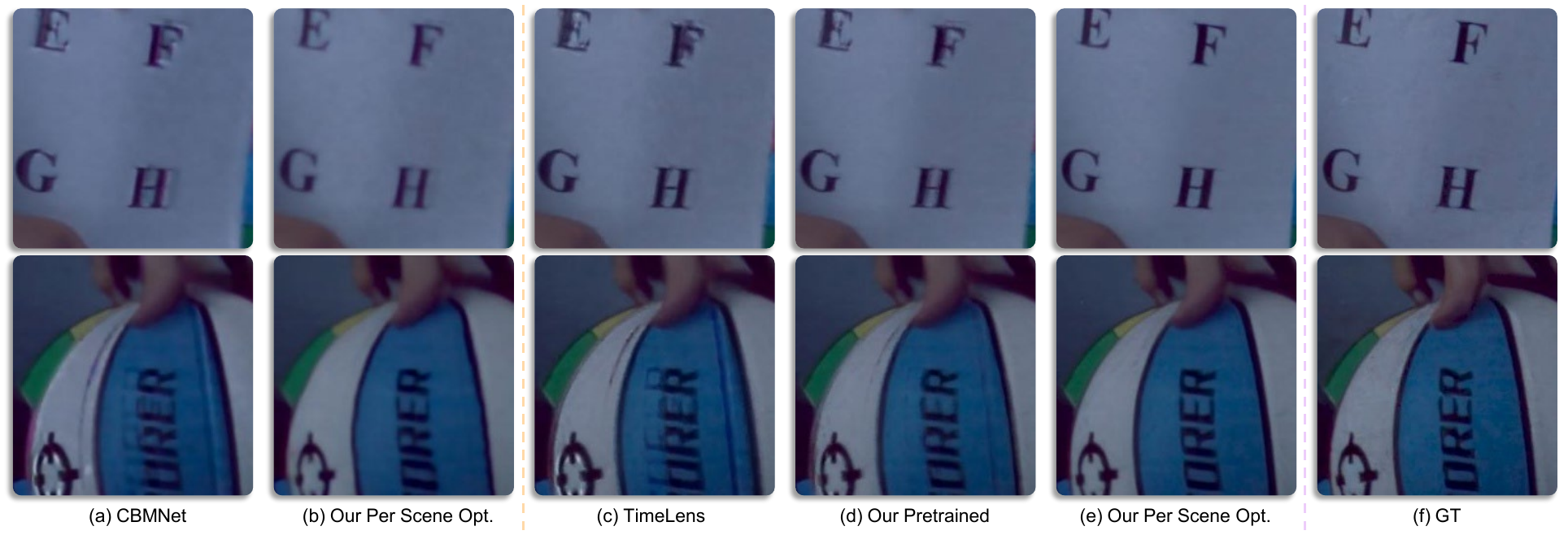}
\caption{Visual comparison of ablation study results. }
\label{fig:ablation}
\end{figure*}

\subsection{Evaluation on Real-world Low-Light Event-VFI}
We evaluated methods on EVFI-LL dataset at 4$\times$ and 8$\times$ interpolation rates across various event triggering thresholds. The quantitative results are presented in Tab.~\ref{tab:bia}. For evaluation, we employed both numerical metrics, such as PSNR and SSIM, and the perceptual metric, LPIPS~\cite{zhang2018perceptual}. We also conducted evaluations on the challenging track, EVFI-LL-C, with quantitative comparisons detailed in Tab.~\ref{tab:high_speed}. Visual comparisons are shown in Fig.~\ref{fig:large_motion}.

In Tabs.~\ref{tab:bia} and \ref{tab:high_speed}, one interesting observation is that when the interpolated ratio is small ($4\times$), event-based methods (\emph{i.e.}, TimeLens and CBMNet) even perform worse than RGB-only methods (\emph{i.e.}, SuperSloMo and RIFE). This is because the degradation of the low-light event signal and the domain gap between simulated event and real-world event significantly hurt the performance of event-based solution. Still, in more challenging $8\times$ interpolation, the Event-VFI methods performs better than RGB-VFI methods, because the additional event stream does provide more information about intermediate motion. Despite this, as shown in Fig.~\ref{fig:large_motion}, the previous methods fail to account for the trailing artifacts of low-light events, leading to visually unpleasing interpolated results with noticeable deformations around object edges.

In contrast, benefiting from per-scene optimization, our enhanced model, Our Per Scene Opt., consistently outperforms the previous methods across all interpolation scales and bias settings. Notably, our model achieves nearly a \emph{2dB} improvement on the entire EVFI-LL dataset compared to the second-best methods. As depicted in Fig.~\ref{fig:large_motion}, our per-scene optimized results preserve crisp details and textures while accurately rendering motions.

\begin{table}[!t]
\centering
\caption{Comparison of Different Frame Interpolation Methods in challenging track in EVFI-LL-C with 8$\times$ Interpolation Rates.}
\begin{tabular}{@{}>{\centering\arraybackslash}p{4.2cm}cccc@{}}
\toprule
\textbf{Method} & \textbf{PSNR $\uparrow$} & \textbf{SSIM $\uparrow$} & \textbf{LPIPS $\downarrow$}\\ 
\midrule
SuperSloMo~\cite{jiang2018super} & 26.269 & 0.8039 & 0.2341  \\
RIFE~\cite{huang2022real} & 26.668 & 0.8122 & 0.2437  \\
TimeLens~\cite{tulyakov2021time} & 30.945 & 0.8401 & 0.1818  \\
CBMNet~\cite{kim2023event} & 30.552 & 0.8431 & 0.2357  \\
\cmidrule(lr){1-4}
Our Pretrained & 31.471 & 0.8543 & 0.1532  \\
Our Per Scene Opt. & \textbf{31.948} & \textbf{0.8695} & \textbf{0.1472}  \\
\bottomrule
\end{tabular}
\label{tab:high_speed}
\end{table}

\subsection{Discussion}
\paragraph{Low light simulation pipeline.}
In Tab.~\ref{tab:ablation-study}, it is evident that Our Pretrained (TimeLens) achieves an improvement of 0.5 to 1 dB over the original TimeLens under low-light conditions, demonstrating the effectiveness of our simulation pipeline.
Fig.~\ref{fig:tail} offers visual comparisons of our simulation with typical v2e simulation. The visualizations distinctly show the sparse and asymmetric event patterns, which are characteristic of low-light conditions and the fundamental operating principles of event cameras. Unlike merely adjusting the parameters of the v2e simulator, our method synthesizes event representations that convincingly mimic real sensor data, including accurate depictions of the elongated event streaks caused by latency. 

\paragraph{Per-scene optimization.}
We propose a per-scene optimization strategy to address the challenges of Event-VFI under low-light conditions. 
By creating supervision pairs directly from the test sequence, the Event-VFI network can be optimized to incrementally adapt to the current scene's specific event degradation. Our experimental results, detailed in Tab.~\ref{tab:ablation-study}, show that our per-scene optimization strategy, \textit{Our Per Scene Opt. (TimeLens)}, significantly enhances interpolation quality under real-world low-light conditions compared to the unoptimized model, \textit{Our Pretrained (TimeLens)}. As illustrated in Fig.~\ref{fig:teaser}, the optimized outcomes reveal sharply recovered motion details, and observe no ghosting or trailing artifacts.

\paragraph{Running time of per-scene optimization.}
We conducted a comparative analysis of the running times for the per-scene optimization stage and the inference stage on an NVIDIA RTX 4090. Given that the optimization process encompasses data loading durations, we calculated the total runtime of the whole program. For the configuration where eight frames are interpolated between two images of size $736 \times 576$, TimeLens exhibited an approximate average runtime of 0.16 seconds, whereas the per-scene optimization process required about average 0.22 seconds. Notably, even though our method incorporates a training component, the time taken for per-scene optimization is significantly less than the time required for training from scratch, which takes approximately three days to train models like  Our Pretrained (TimeLens). The duration of the per-scene optimization is roughly on the same order of magnitude as the inference time. 

\begin{table}[t!]
\centering
\caption{Ablation Study on 8$\times$ Frame Interpolation on EVFI-LL}
\label{tab:ablation-study}
\begin{tabular}{@{}>{\centering\arraybackslash}p{4.2cm}cccc@{}}
\toprule
\textbf{Method} & \textbf{PSNR $\uparrow$} & \textbf{SSIM $\uparrow$} & \textbf{LPIPS $\downarrow$}\\ 
\midrule
CBMNet & 29.265 &0.8509 & 0.2155\\
Our Per Scene Opt. (CBMNet) &31.012 &0.8899 & 0.2101\\
\midrule
TimeLens              & 29.873 & 0.8588 & 0.1774  \\
Our Pretrained (TimeLens)              & 30.343 & 0.8754 & 0.1548 \\
Our Per Scene Opt. (TimeLens)               & 31.808 & 0.8862 & 0.1256 \\
\bottomrule
\end{tabular}
\end{table}

\paragraph{Generalize to other backbones.}
To demonstrate the generalizability of per-scene optimization across different backbones, we also applied such strategy to CBMNet. As shown in Tab.~\ref{tab:ablation-study}, Our Per Scene Opt. (CBMNet) exhibited a performance enhancement of $\sim$1.8 dB in low-light conditions. The visual comparison in Fig.~\ref{fig:ablation} evident that the Our Per Scene Opt. (CBMNet) effectively reduces the impact of event trailing artifacts, yielding visually pleasing results. This improvement underscores the potential of per-scene optimization to significantly enhance the generalizability of other Event-VFI algorithms. Additionally, since the original CBMNet model was not trained considering low-light trailing effects, the performance of Our Per Scene Opt. (CBMNet) were somewhat inferior to those achieved with Our Per Scene Opt. (TimeLens), highlighting the importance of a well-pretrained model.

\paragraph{Non-reference evaluation.}
Due to low-light conditions and camera limitations, capturing noise-free ground truth images is difficult. However, despite the denoising process, the positions of contents in the "ground truth" images (used in Tab.~\ref{tab:bia}) remain accurate. To further minimize the denoising algorithm's impact on evaluation, we used non-reference image quality metrics like MUSIQ~\cite{ke2021musiq} in Tab.~\ref{tab:non-refer}. Our interpolation results, with fewer artifacts, achieved superior MUSIQ evaluations.

\begin{table}[h!]
\centering
\caption{Non-reference Quality Evaluation on the EVFI-LL-C Dataset}
\label{tab:non-refer}
\begin{tabular}{@{}>{\centering\arraybackslash}p{1.5cm}cccc@{}}
\toprule
\textbf{Method} & RIFE & TimeLens & CBMNet & Our Per Scene Opt.\\ 
\midrule
MUSIQ($\uparrow$) &29.71 &31.57 &29.89 &\textbf{32.01}\\
\bottomrule
\end{tabular}
\end{table}
\section{Conclusion}

In conclusion, this study introduced a novel approach to address the challenges inherent in event-based video frame interpolation (Event-VFI) under low-light conditions. Trailing artifacts and other signal degradations present in low-light event data are difficult to accurately model during the training phase. To tackle these issues, we have developed a per-scene optimization strategy tailored for general Event-VFI applications in low-light environments. Empirical evidence demonstrates that our per-scene optimization significantly enhances the generalizability of Event-VFI algorithms in low-light scenarios. This optimization effectively mitigates artifacts caused by event trailing and reduces errors in the interpolated frame positions. Furthermore, to evaluate Event-VFI methods in low-light condition, this study collected the EVFI-LL dataset, specifically tailored for low-light environments and varying event camera settings, further underscores our contribution by providing a robust platform for testing and benchmarking low light Event-VFI algorithms.

\paragraph{Limitations.}
The proposed method currently also has several limitations, which we aim to address in future work. First, the current optimization approach does not account for the impact of blur. Although the existing Event-VFI datasets significantly reduce blur by lowering exposure times, this constraint limits the applicability of the algorithm in broader scenarios. Second, similar to previous works, our datasets have been collected using only a single model of event camera. In the future, we hope to further validate the generalizability of our algorithm across different devices.

\begin{acks}
This research was supported by the Shanghai Artificial Intelligence Laboratory. The work was primarily conducted during Ziran Zhang's internship at the laboratory.
\end{acks}

\bibliographystyle{ACM-Reference-Format}
\bibliography{references}

\end{document}